\pdfoutput=1

\documentclass[11pt]{article}

\usepackage[final]{acl}

\usepackage{times}
\usepackage[T1]{fontenc}
\usepackage[utf8]{inputenc}
\usepackage{latexsym}
\usepackage{amsmath}
\usepackage[bb=dsserif]{mathalpha}
\usepackage{microtype}
\usepackage{inconsolata}
\usepackage{fontawesome5}
\usepackage{graphicx}
\usepackage{soul}
\usepackage{scalefnt}
\usepackage{tabularx}
\usepackage{siunitx}
\usepackage{booktabs}
\usepackage[table,dvipsnames]{xcolor}
\usepackage{blindtext}
\usepackage{enumitem}
\usepackage[most]{tcolorbox}
\usepackage{caption}
\usepackage{xparse}
\usepackage{etoolbox}
\usepackage[capitalize,noabbrev,nameinlink]{cleveref}
\usepackage{mdframed}
\makeatletter
\AddToHook{cmd/appendix/before}{\def\cref@section@alias{appendix}}
\makeatother
\newlength{\currentparskip}%

\newrobustcmd\B{\DeclareFontSeriesDefault[rm]{bf}{b}\bfseries}
\newcommand\smaller[2][0.85]{{\scalefont{#1}#2}}

\lstdefinelanguage{SimpleLang}{
    morecomment=[l]{//}, 
}

\tcbset {
  base/.style={
    arc=0mm,
    bottomtitle=0.5mm,
    boxrule=0mm,
    colbacktitle=black!10!white,
    colframe=black!30!white,
    coltitle=black,
    fonttitle=\bfseries,
    left=2.5mm,
    leftrule=1mm,
    right=3.5mm,
    title={#1},
    toptitle=0.75mm,
    breakable,
    listing options={style=tcblatex, language=SimpleLang, commentstyle=\color{gray}, breakindent=0pt, breaklines=true, basicstyle=\ttfamily\small}
  }
}

\AtBeginDocument{
  \newtcblisting[use counter=lstlisting,number within=chapter,blend into=listings]{codebox}[2][]{
    base={#2},
    listing only,
    #1
  }

  \newtcbinputlisting[use counter=lstlisting,number within=chapter,blend into=listings]{\codeboxinput}[3][]{%
    listing file={#3},
    base={#2},
    listing only,
    #1
  }
}

\newcommand{\circledtext}[1]{%
  \tikz[baseline=(char.base)]{
    \node[
      draw,
      circle,
      inner sep=0.8pt,
      fill=white,
      text=black
    ] (char) {\small #1};
  }%
}

\title{Marcel: A Lightweight and Open-Source Conversational\\ Agent for University Student Support}

\author{Jan Trienes \quad Anastasiia Derzhanskaia \quad Roland Schwarzkopf \\
{\bf Markus Mühling} \quad {\bf Jörg Schlötterer} \quad {\bf Christin Seifert} \\
Marburg University\\
\texttt{\normalsize \{\href{mailto:jan.trienes@uni-marburg.de}{jan.trienes},\href{mailto:christin.seifert@uni-marburg.de}{christin.seifert}\}@uni-marburg.de}}

\begin{document}
\maketitle

\begin{abstract}
We present \emph{Marcel}, a lightweight and open-source conversational agent designed to support prospective students with admission-related inquiries.
The system aims to provide fast and personalized responses, while reducing workload of university staff.
We employ retrieval-augmented generation to ground answers in university resources and to provide users with verifiable, contextually relevant information.
We introduce a Frequently Asked Question~(FAQ) retriever that maps user questions to knowledge-base entries, which allows administrators to steer
retrieval, and improves over standard dense/hybrid retrieval strategies.
The system is engineered for easy deployment in resource-constrained academic settings.
We detail the system architecture, provide a technical evaluation of its components, and report insights from a real-world deployment.

{
    \small
    \makebox[1em][l]{\faGithub}\hspace{0.6em}  \href{https://github.com/aix-group/marcel-chat}{\texttt{github.com/aix-group/marcel-chat}} \\
    \makebox[1em][l]{\faVideo}\hspace{0.6em} \href{https://youtu.be/uLCB2R6szz4}{\texttt{youtu.be/uLCB2R6szz4}}
}
\end{abstract}

\section{Introduction}
Prospective university students often face challenges navigating the admission requirements.
While many universities are investing in student support services, these typically do not grow at the same rate as many study programs.
To contextualize this problem in a real-world example, we consider the M.Sc Data Science program at our institution.
The growing interest in this subject and a recent internationalization of the program lead to a stark increase in applications.
However, many applications had to be rejected because formal criteria or prerequisites were not met.
This substantially increased the workload of university staff and led to missed opportunities for students.
Expanding support services with alternative information channels appears necessary to better inform applicants and to reduce the workload of university staff.

A potential solution to this problem are conversational search systems like chatbots~\cite{Zamani:2023:FTIR,Mo:2025:arXiv}.
We posit that these systems make it easier for students to find and understand relevant information due to their interactivity.
Recent advances in Retrieval Augmented Generation (RAG, \citealp{Lewis:2020:NeurIPS}) allow language models to provide grounded and verifiable answers based on university resources.
However, operating such a RAG-based system in a university setting is challenging.
Many universities require on-premise deployments to comply with privacy laws, have limited computational resources, and need cost-effective systems that are easy to customize, maintain and monitor.
Several universities are actively developing or using such chatbots~\cite{Nguyen:2025:URAG,Odede:2024:CHIIR,Cherumanal:2024:CHIIR,Brereton:2021:Online} which indicates high demand.
However, they are typically closed-source which limits others to build on those efforts.

\begin{figure*}[t]
    \includegraphics[width=\textwidth]{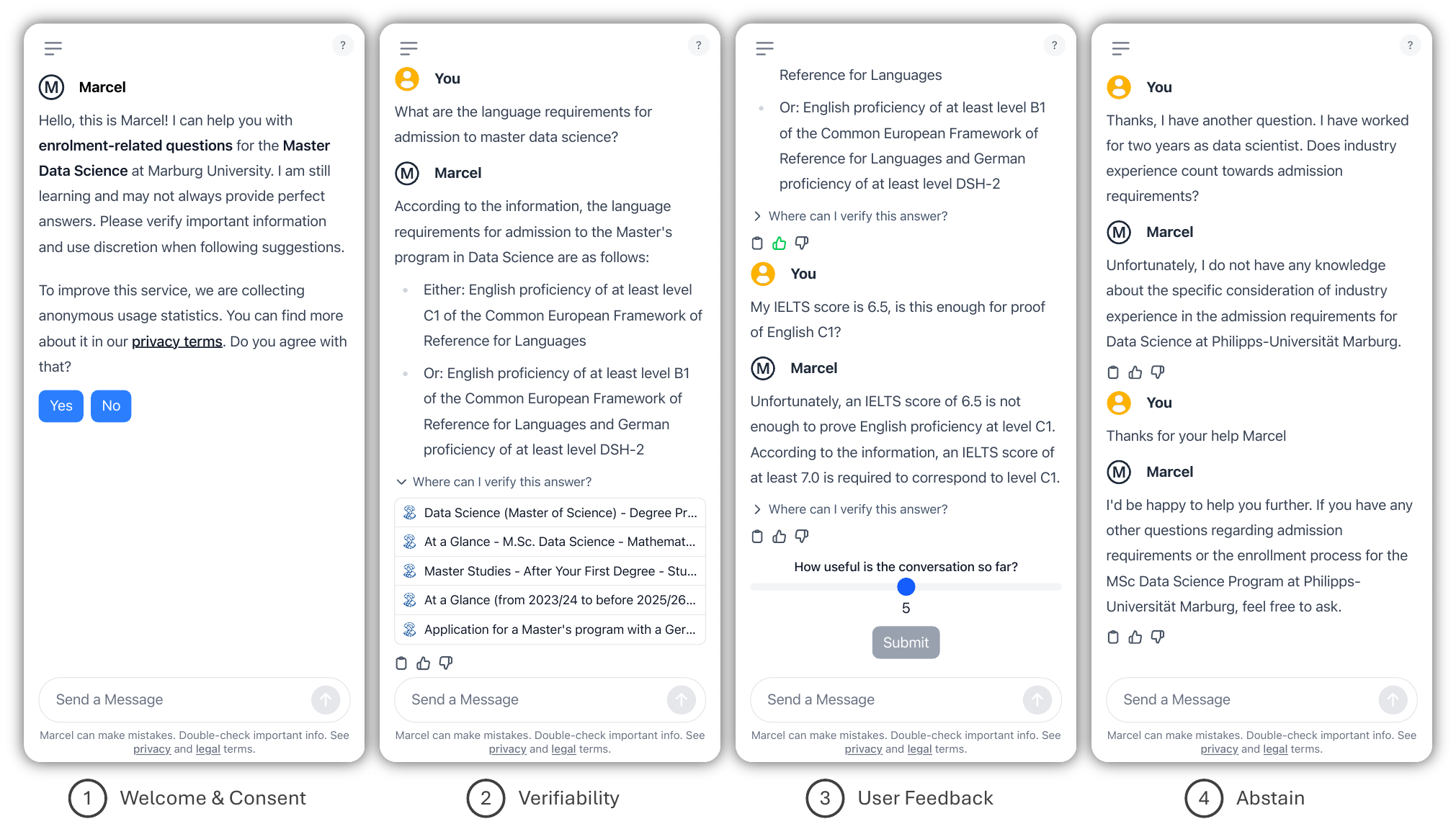}
    \caption{Overview of student-facing UI. \protect\circledtext{1} Welcome and consent are shown on first use. \protect\circledtext{2} For each generated answer, Marcel provides the list of source documents which are linked to the respective websites. \protect\circledtext{3} Users can rate individual responses using a thumbs up/down button. The overall conversation can be rated after three interactions on a 10-point Likert scale. \protect\circledtext{4} Marcel abstains from answering questions that are not reflected in the underlying knowledge base. In those cases, no list of source documents is shown. An admin-facing UI is given in \cref{fig:admin-interface}.}
    \label{fig:user-interface}
\end{figure*}

To bridge this gap, we introduce \emph{Marcel} a student chatbot aimed at answering enrollment-related questions.\footnote{For this demo, we focus on enrollment-related questions for the M.Sc Data Science program at our institution. However, all design decisions are made to generalize to other information-seeking tasks in the university setting.}
This demo makes three contributions:

\textbf{(1) FAQ Retriever} (\cref{sec:system-description}).
We introduce a novel retriever which maps user queries to a set of curated frequently asked questions with associated answer-relevant documents.
This component allows administrators to steer retrieval, and improves over dense/hybrid retrieval strategies.

\textbf{(2) Efficient Deployment} (\cref{sec:experiments}).
We outline our design decisions to facilitate deployments with minimal infrastructure (i.e., a single node with a GPU).
We empirically evaluate several retrieval and generation strategies to better understand quality-efficiency tradeoffs in this task.

\textbf{(3) Practical Insights} (\cref{sec:deployment-insights}).
We share qualitative insights from deploying the chatbot in practice.
To support adoption, we release our system as containerized software along with supporting artifacts such as content scraper and deployment scripts to enable customization.\footnote{All software is released under the MIT license.}

\section{System Description}
\label{sec:system-description}

\subsection{Question Answering}
\label{sec:faq-retriever}
\textbf{Document retrieval.}
\emph{Marcel} uses RAG to provide answers grounded in university resources.
To improve retrieval precision and add a layer of administrative control, we introduce an FAQ retriever that maps queries to a set of curated questions, each linked to one or more relevant documents.
The closest FAQ is selected through semantic similarity of question embeddings. Intuitively, FAQ retrieval acts as a lightweight intent classifier.
We combine it with standard retrievers (e.g., BM25) to provide fallbacks for question intents falling outside the FAQ.
Although FAQ curation requires manual effort, many universities maintain FAQ-like resources (e.g., for student support services) that can be bootstrapped. For our deployment, we worked with university staff to establish 36 FAQs, reflecting most enrollment related information-needs.
We provide example FAQs in Appendix, \cref{tab:faq-example}.

\textbf{Answer generation.}
An instruction-tuned Large Language Model (LLM) generates answers conditioned on retrieved documents.
To improve answer grounding and dialogue handling,  answering is done in two stages.
First, an LLM-based classifier determines whether a query requires retrieval or is chit-chat or a meta-question (e.g., ``What topics can you help with?'').
If so, the system responds without retrieval.
Otherwise, the LLM is instructed to answer only if the retrieved documents provide sufficient information, or to abstain and return a fallback response (e.g., ``Unfortunately, I cannot answer based on...'').
See \cref{sec:appendix-prompts} for the prompts of query classifier and answer generator.

\textbf{Knowledge base.}
We scrape enrollment-related documents from university websites (e.g., study descriptions, admission requirements and procedures, course catalog, and accommodation services).
Curating the knowledge base through scraping is efficient, since latest information is published on the websites.
Periodically re-running the scraper captures changes automatically (e.g., updated admission requirements).
Pre-processing of scraped content is kept minimal: to reduce token count in downstream generation, we convert HTML to Markdown and replace links with numeric IDs.
We avoid splitting documents to preserve context.
The corpus consists of 249 documents, with an average length of 712 tokens using the \textsc{nltk} tokenizer.

\subsection{User Interaction}
We support two user groups: (i) students seeking information, and (ii) administrators wanting to monitor and evaluate the system.
The student-facing UI adopts familiar conventions of chat interfaces and is fully responsive (\cref{fig:user-interface}).
On first visit, a consent dialog informs users about data collection and chatbot limitations.
The disclaimer states that generated answers may be inaccurate and must be verified.
Users need to confirm that they have read this message before continuing.
We discuss broader ethical implications of the chatbot at the end of the paper.

Responses are shown alongside links to retrieved sources to support verification.
Users can rate each response with a thumbs-up/down, and after several turns, provide conversation-level feedback on a Likert scale.
If a query is not answerable with retrieved context, the system abstains and hides sources.
The admin UI shows usage statistics over a user-defined period and allows to review full conversations and ratings (Appendix, \cref{fig:admin-interface}).

\subsection{Software Architecture and Deployment}
Marcel is designed for fast deployment in resource-constrained academic settings.
The chatbot is implemented as a containerized architecture with four components: a JavaScript frontend (Vue.JS), a Python backend (FastAPI), a relational database (MariaDB), and an LLM inference engine (vLLM, \citealp{Kwon:2023:SIGOPS}).
A reverse proxy (nginx) routes traffic and handles HTTPs encryption.
We run Marcel on a node with 2 A100 GPUs for vLLM\footnote{The newer Gemma 27b model runs on a single A100. With 4bit quantization \href{https://developers.googleblog.com/en/gemma-3-quantized-aware-trained-state-of-the-art-ai-to-consumer-gpus}{a 3090 is supposed to be sufficient}.} and a VM (4 cores, 4GB RAM) for the remaining containers.
We implement best practices for container security by \citet{OWASP:2025:Online}, including (i) container isolation, (ii) minimal rights, (iii) not exposing the Docker socket.
Automated, reproducible deployments and updates are done using Ansible.
To support adoption by other institutions, we share both the Ansible scripts and regular docker-compose files as a blueprint.

\section{Technical Evaluation}
\label{sec:experiments}
We conduct a technical evaluation of \emph{Marcel's} core components to better understand the quality and efficiency trade-offs in this resource-constrained university chatbot setting. Our experiments are guided by the following research questions:

\begin{enumerate}[
    label=\textbf{RQ\arabic*.},
    ref={RQ\arabic*},
    labelwidth=3em,
    labelsep=0.5em,
    leftmargin=*,
    itemindent=0pt,
    align=parleft,
    listparindent=0pt,
    noitemsep
    ]
    \item How does the FAQ-augmented retriever compare to standard retrieval strategies?
    \item To what extent can a second-stage reranker improve retrieval quality?
    \item How well do small-scale, open-weights generators perform in this domain?
    \item What is the impact of quantization on generation quality and efficiency?
\end{enumerate}

\subsection{Experimental Systems}

\paragraph{Retrievers.}
To address RQ1 and RQ2, we evaluate diverse retrieval strategies commonly used in production systems, spanning lexical, dense, and hybrid methods, with and without reranking.
\begin{itemize}[noitemsep]
    \item \textbf{BM25}~\cite{Robertson:2009:FTIR}. Standard lexical search, which is a strong baseline in practical settings.
    \item \textbf{Dense}. Document relevance is determined by the similarity between query and document embeddings. Document embeddings are precomputed during indexing using models optimized for semantic search in sentence-transformers~\cite{Reimers:2019:EMNLP}. Specifically: \href{https://huggingface.co/sentence-transformers/all-MiniLM-L6-v2}{\texttt{\small MiniLM}} (22.7M params) and \href{https://huggingface.co/sentence-transformers/msmarco-bert-base-dot-v5}{\small\texttt{MS MARCO Bert}} (109M params).
    \item \textbf{HyDE}~\cite{Gao:2023:ACL}. An LLM is prompted to generate $n$ artificial documents that could plausibly answer a given query. The average embedding of these documents is used as the query, similar to dense retrieval. To understand how generator scale affects HyDE performance, we test the Gemma 3 model family~\cite{Gemma:2025:arXiv}. We set $n=3$ and use the same embedding model as the dense retriever (\textsc{ms marco}).
    \item \textbf{FAQ}. Retrieving relevant documents via FAQ matching (see \cref{sec:faq-retriever}).
\end{itemize}
We evaluate combinations of above retrievers by merging document rankings through reciprocal rank fusion~\cite{Cormack:2009:SIGIR}.
To complement these retrievers, we evaluate second-stage reranking using cross-encoders (\textbf{Rerank}).\footnote{We use \href{https://huggingface.co/mixedbread-ai/mxbai-rerank-base-v1}{\texttt{mxbai-rerank-base-v1}}, which performs well on general benchmarks~\cite{Shakir:2024:Blog}; See \cref{appendix:reranker-evaluation} for an evaluation of other representative models in the 33M to 435M parameter range.}
For retrieval evaluation, we take the top $k=50$ documents. For generation, we restrict to the top $k=5$ documents to meet latency and context length constraints.

\paragraph{Generators.}
We seek a generator that is (i) open-weight, allowing for a local and privacy-compliant operation, and (ii) that is deployable in hardware constrained settings (e.g., 1--2 GPUs). The current deployment of \emph{Marcel} uses Llama 3.1~\cite{Grattafiori:2024:arXiv}, in the 70B variant with int8 quantization. For comparison, we benchmark a second candidate, Gemma 3~\cite{Gemma:2025:arXiv}, because of its strong performance in the Chatbot Arena~\cite{Chiang:2024:ICML}.\footnote{At the time of study (June 2025), Gemma 3 was the highest ranked model that can be deployed on a single GPU (ranked 28/254). The next better open-weights model was Qwen 3 with 235B parameters (ranked 23/254).}
To address RQ3 and RQ4, we evaluate all available model sizes for both Llama and Gemma families, and assess the impact of int8 quantization for Llama models.
As retriever we use BM25+FAQ. Additionally, we use an Oracle to estimate generation quality under optimal retrieval.
For each query, we generate three outputs with temperature sampling ($\tau = 0.7$), and report evaluation metrics averaged over these generations. The system prompt is provided in Appendix,~\cref{lst:generator-prompt}.

\subsection{Evaluation Protocol}

\paragraph{Data.}
In the initial development phase of \emph{Marcel}, we lacked real user queries directed at the chatbot.
Therefore, we bootstrapped an evaluation dataset from student emails sent to university support offices.
This provides a realistic approximation of the information needs that end-users have.
To simplify the analysis, we focus on single-turn interactions.

Emails were collected and anonymized during a time frame of 6 months. We performed light de-duplication of overrepresented information needs.
Each question is associated with a manually written ground-truth response and the relevant sources in the knowledge base (i.e., document IDs).
We distinguish between questions that are \emph{answerable} and \emph{unanswerable} with the knowledge base, to assess the generator's ability to abstain when the retrieved context does not provide relevant information (e.g., questions about visa issues, late arrivals, conditional accepts, etc.).
See \cref{tab:evaluation-dataset} for an overview of the dataset and Appendix, \cref{tab:query-examples} for examples.

\begin{table}[t]
\small
\centering
\resizebox{\linewidth}{!}{%
\begin{tabular}{lrrr}
\toprule
\textbf{Statistic} & \textbf{All} & \textbf{Answerable} & \textbf{Unanswerable} \\
\midrule
Questions & 95 & 76 & 19 \\
Question length & 43.3 & 40.5 & 54.3 \\
Answer length & 45.5 & 46.0 & 43.6 \\
Answer sources & 1.0 & 1.3 & 0.0 \\
FAQs & 36 & 25 & 11 \\
FAQ length & 9.7 & 8.9 & 12.8 \\
\bottomrule
\end{tabular}}
\caption{Summary statistics of the evaluation dataset.}
\label{tab:evaluation-dataset}
\end{table}

\paragraph{Metrics.}
We evaluate retrieval by Mean Reciprocal Rank (\textbf{MRR}) and Recall at Cutoff (\textbf{R@K}).
To evaluate generation, we employ both reference-based and reference-free metrics, using LLM-as-a-Judge.
For reference-based metrics, we report \textbf{ROUGE-1}~\cite{Lin:2004:WS} and \textbf{BERTScore}~\cite{Zhang:2020:ICLR}.
For reference-free evaluation, we report \textbf{Answer Faithfulness} and \textbf{Answer Relevance}~\cite{Es:2024:EACL}.
Answer faithfulness estimates the fraction of answer claims supported by the retrieved context.
Answer relevance generates $n$ hypothetical questions from the answer, and measures their similarity to the original query.

Additionally, we evaluate the generator's ability to abstain when a query cannot be answered with the knowledge base.
Abstaining is important, as any answer generated from the parametric knowledge of the LLM is not grounded in the retrieved context and may therefore be incorrect, irrelevant or outdated.
To this end, we define \textbf{Selectivity}.
Let $a_{i,j} \in \{0, 1\}$ indicate whether generation $j$ is an attempt at answering query $i$, and let $y_{i} \in \{0, 1\}$ denote the ground-truth obtained from manual annotation.
Given $N$ queries and $M$ generations,
\begin{equation*}
    \text{Selectivity} = \frac{1}{N}\sum^N_{i=1} \left(\frac{1}{M}\sum^{M}_{j=1}\mathbb{1}[a_{i,j}=y_i]\right),
\end{equation*}
where $\mathbb{1}$ is the indicator function.
Selectivity ranges from 0 to 1, where higher is better.
To obtain $a_{i,j}$, we instruct an LLM to detect hedging phrases (e.g., \emph{I don't know}, \emph{The documents do not provide any information}).
See prompt in Appendix, \cref{lst:selectivity-prompt}.
We report all generator metrics separately for \emph{answerable} and \emph{unanswerable} queries.
This allows us to understand if the system abstains correctly when questions cannot be answered.

Finally, we report computational efficiency as \textbf{seconds per query}. Experiments were conducted on a Slurm node with four CPU cores and two A100 GPUs (80GB), with timings collected after pipeline warmup. All LLM-as-a-Judge metrics are computed using Llama 3.1 70B (int8 quantized).\footnote{While Llama could be favoring its own generations, \citet{Panickssery:2024:NeurIPS} showed that this mainly applies to pairwise evaluations, not direct ratings used in our experiments.}

\subsection{Results}
We present the evaluation of retrievers in \cref{tab:retriever-evaluation} and the evaluation of generators in \cref{tab:generator-evaluation}.

\begin{table}[t]
\sisetup{
detect-weight,
mode=text,
table-format=-1.2,
table-space-text-post={$^{*}$}
}
\small
\setlength{\tabcolsep}{1pt}
\resizebox{\linewidth}{!}{
\begin{tabular}{lSSSSSS}
\toprule
\textbf{Retriever} & \bfseries \text{MRR} & \bfseries \text{R@1} & \bfseries \text{R@5} & \bfseries \text{R@50} & \bfseries \text{sec/q} & \bfseries $\text{sec/q}^\dagger$ \\
\midrule
\multicolumn{7}{l}{\emph{Baseline}} \\\midrule
BM25 & 0.26 & 0.09 & 0.37 & 0.74 & \bfseries 0.02 & \color{gray} \text{---} \\
+ Dense (\textsc{minilm}) & 0.31 & 0.15 & 0.44 & 0.82 & 0.06 & 0.04 \\
+ Dense (\textsc{marco}) & 0.40 & 0.21 & 0.50 & 0.78 & 0.23 & 0.06 \\\midrule
\multicolumn{7}{l}{\emph{Hyde}~\cite{Gao:2023:ACL}} \\\midrule
+ HyDE (G-1B) & 0.28 & 0.13 & 0.34 & 0.73 & \color{gray} \text{---} & 4.81 \\
+ HyDE (G-4B) & 0.38 & 0.25 & 0.41 & 0.82 & \color{gray} \text{---} & 5.22 \\
+ HyDE (G-27B) & 0.44 & 0.29 & 0.49 & \bfseries 0.84 & \color{gray} \text{---} & 20.92 \\\midrule
\multicolumn{7}{l}{\emph{FAQ Retriever (Ours)}} \\\midrule
+ FAQ (\textsc{minilm}) & \bfseries 0.70$^{*}$ & \bfseries 0.57$^{*}$ & \bfseries 0.77$^{*}$ & \bfseries 0.84 & 0.05 & \bfseries 0.03 \\
+ FAQ (\textsc{marco}) & 0.61$^{*}$ & 0.47$^{*}$ & 0.63 & 0.80 & 0.19 & 0.03 \\\midrule
\multicolumn{7}{l}{\emph{Reranker} (\texttt{mxbai-rerank-base-v1})} \\\midrule
+ Dense + Rerank & 0.46 & 0.33 & 0.48 & 0.78 & \color{gray} \text{---} & 0.59 \\
+ HyDE + Rerank & 0.53 & 0.38 & 0.55 & 0.83 & \color{gray} \text{---} & 21.41 \\
+ FAQ + Rerank & 0.48 & 0.34 & 0.49 & \bfseries 0.84 & \color{gray} \text{---} & 0.56 \\
\bottomrule
\end{tabular}}
\caption{Evaluation of retrieval strategies. Statistically significant differences relative to baseline (\smaller{BM25 + DENSE MARCO}) are denoted by $^{*}$ at a significance level of $\alpha = 0.01/40$ (Bonferroni correction). Seconds per query are reported on CPU (sec/q) and GPU (sec/q$^\dagger$), while HyDE and Rerank are only tractable on GPU.
}
\label{tab:retriever-evaluation}
\end{table}

\paragraph{Effectiveness of FAQ retriever (RQ1).}
The FAQ retriever significantly outperforms standard retrieval methods, achieving a 75\% increase in MRR over the BM25+Dense baseline. 
This gain in early-rank retrieval quality is particularly valuable for downstream generation, which typically benefits from smaller context sizes both in output quality and throughput.
Importantly, strong performance is achieved even with lightweight embeddings (MiniLM, 22.7M parameters), enabling fast, CPU-friendly retrieval with low latency (0.05 seconds).
Overall, these findings suggest that the FAQ retriever is an effective mechanism for administrators to steer retrieval, with the trade-off of manual FAQ curation and maintenance.

\paragraph{Effectiveness of reranking (RQ2).}
We next examine the impact of second-stage reranking on retrieval quality.
While both BM25+Dense and BM25+HyDE retrieval show modest improvements at early ranks, the quality of FAQ retrieval is substantially degraded ($\approx31\%$ drop in MRR). 
This suggests that current rerankers may not generalize to this domain.
One potential mitigation strategy is to exclude FAQ-retrieved documents from reranking.
However, this approach is only feasible when the FAQ retriever reliably returns relevant documents.
Future work could investigate techniques to dynamically select the most appropriate retriever per query, or to fine-tune rerankers.

\begin{table*}[t]
\small
\centering
\setlength{\tabcolsep}{3pt}
\begin{tabular}{lp{8pt}rrrrrrp{8pt}rSc}
\toprule
\textbf{Generator} & & \textbf{ROUGE} & \textbf{BERTScore} & \textbf{Faithfulness} & \textbf{Relevance} & \textbf{Selectivity} & \textbf{Average} & & \textbf{|ŷ|} & \textbf{sec/q} & \textbf{GPUs} \\
\midrule
\multicolumn{12}{c}{\texttt{Answerable with knowledge base}}\\
\midrule
Gemma 3 (1B) & & 0.17 & -0.03 & 0.33 & 0.73 & 0.38 & 0.32 & & 142 & 7.2$^\dagger$ & 1 \\
Gemma 3 (4B) & & 0.24 & 0.04 & 0.56 & 0.72 & \bfseries 0.88 & 0.49 & & 206 & 3.3 & 1 \\
Gemma 3 (12B) & & 0.30 & 0.15 & \bfseries 0.76 & 0.71 & 0.79 & 0.54 & & 113 & 4.1 & 1 \\
Gemma 3 (27B) & & \bfseries 0.31 & \bfseries 0.16 & 0.71 & 0.70 & \bfseries 0.88 & \bfseries 0.55 & & 114 & 8.0 & 1 \\
\color{gray} $\rightarrow$ Oracle retriever & & \color{gray} 0.35 & \color{gray} 0.18 & \color{gray} 0.63 & \color{gray} 0.68 & \color{gray} 0.91 & \color{gray} 0.55 & & \color{gray} 126 & \color{gray} 6.8 & \color{gray} 1 \\ \midrule
Llama 3.1 (8B) &  & 0.22 & 0.10 & 0.54 & 0.74 & 0.44 & 0.41 & & 89 & 2.4 & 1 \\
$\rightarrow$ int8 quantized & & 0.20 & 0.07 & 0.54 & \bfseries 0.75 & 0.43 & 0.40 & & 93 & 1.9 & 1 \\
Llama 3.1 (70B) & & 0.29 & 0.16 & 0.70 & 0.74 & 0.76 & 0.53 & & 132 & 12.5 & 2 \\
$\rightarrow$ int8 quantized & & 0.29 & 0.16 & 0.73 & 0.73 & 0.80 & 0.54 & & 130 & 8.3 & 2 \\
\midrule
\multicolumn{12}{c}{\texttt{Not Answerable with knowledge base}}\\
\midrule
Gemma 3 (1B) & & 0.16 & 0.02 & 0.31 & 0.64 & 0.65 & 0.32 & & 181 & 7.1$^\dagger$ & 1 \\
Gemma 3 (4B) & & 0.16 & 0.04 & 0.54 & 0.61 & 0.56 & 0.49 & & 161 & 2.9 & 1 \\
Gemma 3 (12B) & & 0.21 & \bfseries 0.14 & 0.70 & 0.56 & 0.75 & 0.54 & & 59 & 2.7 & 1 \\
Gemma 3 (27B) & & \bfseries 0.21 & 0.12 & \bfseries 0.75 & 0.60 & 0.68 & \bfseries 0.55 & & 80 & 6.5 & 1 \\
\color{gray} $\rightarrow$ Oracle retriever & & \color{gray} 0.18 & \color{gray} 0.13 & \color{gray} --- & \color{gray} 0.67 & \color{gray} 0.93 & \color{gray} 0.48 & & \color{gray} 43 & \color{gray} 2.1 & \color{gray} 1 \\\midrule
Llama 3.1 (8B) & & 0.15 & 0.13 & 0.47 & 0.55 & \bfseries 0.91 & 0.41 & & 53 & 1.6 & 1 \\
$\rightarrow$ int8 quantized & & 0.17 & 0.14 & 0.52 & 0.55 & 0.89 & 0.40 & & 41 & 1.0 & 1 \\
Llama 3.1 (70B) & & 0.16 & 0.11 & 0.55 & 0.61 & 0.67 & 0.53 & & 64 & 9.5 & 2 \\
$\rightarrow$ int8 quantized & & 0.18 & 0.13 & 0.55 & \bfseries 0.66 & 0.63 & 0.54 & & 67 & 5.5 & 2 \\
\bottomrule
\end{tabular}
\caption{Evaluation of generators stratified by whether a query is answerable with the knowledge base. Additionally to quality metrics, we report the length of response in tokens ($|\hat{y}|$) and processing time per query (sec/q). $^\dagger$ = We were unable to obtain reliable timings for Gemma 3 1B due to a vLLM bug (v0.9.1, \href{https://github.com/vllm-project/vllm/issues/19575}{vllm/\#19575}).}
\label{tab:generator-evaluation}
\end{table*}

\paragraph{Generator evaluation (RQ3).}
We find that Gemma (12B, 27B) and Llama (70B) perform comparably, especially on the \emph{answerable} subset of queries (\cref{tab:generator-evaluation}).
Gemma (12B) strikes a good balance between quality and resource efficiency, showing that it may not be necessary to resort to the largest available models.
We note an interesting model-dependent trend on response length: larger Llama models tend to produce longer responses, while Gemma models show the opposite pattern.
However, we cannot conclude that either behavior leads to higher response quality.

Substituting the FAQ retriever with an oracle retriever results in minimal changes in performance on answerable queries.
This highlights the quality of the FAQ retriever, but also indicates that the current evaluation protocol may be insensitive to improvements beyond a certain threshold.
On unanswerable queries, the use of an oracle retriever significantly increases the rate of correct abstentions (0.68$\rightarrow$0.93), suggesting that more conservative retrieval can further improve faithfulness.
Additionally, it may be worthwhile to investigate alternative strategies for classifying whether or not a question is answerable given the retrieved context~\cite{Peng:2025:ACL}. Currently, this task is left to the answer generation prompt (cf. \cref{sec:faq-retriever}) which may leave room for improvement.

\paragraph{Impact of quantization (RQ4).}
We find that int8 quantization has no noticeable effect on generation quality for both Llama 8B and 70B. At the same time, it significantly reduces inference latency (speedup of $\approx20\%$--$42\%$).
This makes quantization a suitable technique to reduce hardware requirements for practical deployments of \emph{Marcel}.

\section{Insights from End-User Deployment}
\label{sec:deployment-insights}
We deployed Marcel during the final weeks of the current admissions cycle (winter term 2025/26).
Despite limited user engagement due to the timing, this deployment demonstrates production-readiness and allowed us to collect qualitative insights from end-user interactions.
Over a nine-week period, Marcel served 92 users, exchanging 926 messages in 126 conversations, without any system errors or downtime.
Below, we discuss several qualitative observations and examples that reflect current limitations of the chatbot.

\emph{(1) Answers are tailored, but not proactive (Appendix, \cref{fig:example-proactivity}).}
Some information-needs could be addressed more efficiently through proactivity.
For example, instead of answering ``Since only 50\% your BSc courses are related to Data Science, you may not be admitted,'' the chatbot could prompt the user to provide specific course contents, to subsequently match them against admission requirements.
However, these interactions require strong model reasoning and carry a high risk of ungrounded inference.

\emph{(2) Knowledge gaps (Appendix, \cref{fig:example-knowledge-gap}).}
We encountered several questions that the chatbot abstained from answering.
This is caused either by an incomplete knowledge base, or because the website lacks this information.
While frequent abstention limits the chatbot usefulness, it improves grounding, and has the beneficial side-effect of identifying ways to improve clarity and completeness of university websites.

\emph{(3) Contextualized queries (Appendix, \cref{fig:example-contextualization}).}
The chatbot struggles with queries contextualized in multi-turn dialog.
For example, follow-up questions often lack information of previous queries, causing the retriever to surface irrelevant documents.
An adjacent problem are underspecified queries.
For example, many user questions do not explicitly call out that the query is about the data science program (e.g., ``When is the deadline for application?''), which is a consequence of users entering the chatbot through program-specific pages.
Practically, this degrades retrieval quality and can lead to unfocused generations when the retrieved context provides multiple valid or even conflicting answers.
A potential solution could be query expansion via decontextualization~\cite{Choi:2021:TACL}.

\emph{(4) Link-seeking queries (Appendix, \cref{fig:example-link-seeking}).}
While source links are shown alongside answers, the chatbot cannot embed them within responses as they are usually absent from source documents.
To enable link-seeking interactions, it may be necessary to embed links in the documents and to instruct the generator to cite them.

\emph{(5) Little uptake of feedback mechanisms.}
We observed little user engagement with the feedback mechanisms.
The message-level rating (thumbs-up/down) was used only once, and only 11 conversations received a Likert-scale rating.
This highlights a general challenge in obtaining direct user feedback.
Downstream evaluation may therefore need to focus on answer quality.
In addition, latent usage signals such whether users ask followup questions or whether they return to the chatbot could be taken into account.

\section{Related Work}
RAG~\cite{Lewis:2020:NeurIPS} is becoming a widely adopted strategy across many domains in conversational search~\cite{Zamani:2023:FTIR}.
In the educational domain, \citet{Nguyen:2021:Chatbot} deployed a chatbot to support admissions using the dialog management framework Rasa~\cite{Bocklisch:2017:arXiv}. Significant engagement demonstrates high student demand.
\citet{Odede:2024:CHIIR} conducted a user study comparing live-chat with university staff to a RAG-chatbot.
Participants valued fast responses and low-entry barriers of the chatbot.

\citet{Cherumanal:2024:CHIIR} find that intent-based retrieval can be an effective alternative to document retrieval, which supports the overall idea of our FAQ retriever.
Closely related, \citet{Nguyen:2025:URAG} proposed a two-tiered strategy: if a query is sufficiently similar to a pre-defined FAQ, the system returns a pre-defined answer; otherwise, it falls back to document retrieval.
Our system differs in two ways.
First, instead of pre-defining answers, we link FAQs to knowledge base documents to ensure grounded and up-to-date answers.
Second, instead of imposing a minimum similarity between queries and FAQ entries, we use the similarity as a soft ranking signal combined with other retriever scores which avoids an additional hyperparameter.

In the general domain, document assistants such as TruthReader~\cite{Li:2024:EMNLP} and Verba~\cite{Weaviate:2025:Online} follow a similar RAG architecture as our system, but assume user-provided documents.
The RAG-framework \citet{Onyx:2025:Online} provides rich utilities to connect to heterogeneous data sources, but allows little customization of the retrieval strategy.

Despite these advances, there remains a gap in open-source chatbots designed for deployment in resource-constrained university settings, which our paper aims to address.

\section{Conclusion}
We present \emph{Marcel}, an open-source RAG-based chatbot designed for handling enrollment-related student questions.
The design and architecture targets resource-constrained university settings, taking into account privacy, controllability and observability.
We analyze quality-efficiency trade-offs of common RAG-strategies, and provide insights from a real-world deployment at our institution.
We plan to improve answer quality through query expansion and to integrate mechanisms for online evaluation of different answering-strategies.

\section*{Limitations}
We note two limitations of our study.
First, we considered only one prompt for the answer generator.
It is conceivable that different prompting strategies could be used to improve performance, in particular for deciding when to abstain from answering a query.
Second, we have only reported an intrinsic evaluation of the chatbot components.
A longer-term evaluation is necessary to see if the system improves information access for students.
As mentioned in the introduction, we hope that improved information access leads to a reduction in the number of unsuccessful applications due to formal criteria, and subsequently reduces the workload of academic staff.
Understanding whether the chatbot will help to reach those goals requires a long-term evaluation and additional user uptake.

\section*{Ethical Considerations}
It is important to note that university admissions is a high-risk domain, where wrong chatbot responses may adversarially affect users.
At the current stage of the technology, it cannot be guaranteed that the chatbot provides correct answers due to the inherently stochastic nature of LLMs.
Therefore, the chatbot only serves as an \emph{additional} information source next to existing student support structures in form of the website, FAQs, email, and phone.
To mitigate the risks associated with wrong responses, we (i) insert prominent disclaimers in the user interface, (ii) encourage users to verify important information through the display of relevant sources, and (iii) provide an admin interface to allow periodic review of conversations.
A useful extension of this would be automatic flagging of potentially inaccurate responses so that they can be reviewed more efficiently.
Lastly, all users in our study gave their explicit consent for the data collection and can retroactively opt-out.

\bibliography{bibliography}

\clearpage
\onecolumn
\appendix
\section{Benchmarking Rerankers}
\label{appendix:reranker-evaluation}
\begin{table*}[h!]
\small
\setlength{\tabcolsep}{3pt}
\resizebox{\textwidth}{!}{
\begin{tabular}{lrrrrrrrrrrr}
\toprule
\bfseries Retriever & \bfseries MRR & \bfseries $\Delta$MRR & \bfseries R@1 & \bfseries $\Delta$R@1 & \bfseries R@5 & \bfseries $\Delta$R@5 & \bfseries sec/query & \bfseries $\Delta$sec/query & \bfseries Params \\
\midrule
BM25 + Dense (\textsc{marco}) & 0.40 & \color{gray}{---} & 0.21 & \color{gray}{---} & \bfseries 0.50 & \color{gray}{---} & \bfseries 0.06 & \color{gray}{---} & \color{gray}{---} \\\midrule
+ Rerank (\texttt{ms-marco-MiniLM-L12-v2}) & 0.23 & -42\% & 0.08 & -62\% & 0.35 & -30\% & 0.19 & 217\% & 33.4M \\
+ Rerank (\texttt{jina-reranker-v1-turbo-en}) & 0.26 & -35\% & 0.10 & -52\% & 0.40 & -20\% & 0.17 & 183\% & 37.8M \\
+ Rerank (\texttt{ms-marco-MiniLM-L6-v2}) & 0.27 & -32\% & 0.11 & -48\% & 0.36 & -28\% & 0.14 & \bfseries 133\% & 22.7M \\
+ Rerank (\texttt{jina-reranker-v1-tiny-en}) & 0.27 & -32\% & 0.12 & -43\% & 0.37 & -26\% & 0.14 & \bfseries 133\% & 33M \\
+ Rerank (\texttt{mxbai-rerank-large-v1}) & 0.37 & -8\% & 0.20 & -5\% & 0.45 & -10\% & 1.59 & 2550\% & 435M \\
+ Rerank (\texttt{mxbai-rerank-xsmall-v1}) & 0.42 & 5\% & 0.30 & 43\% & 0.46 & -8\% & 0.30 & 400\% & 70.8M \\
+ Rerank (\texttt{mxbai-rerank-base-v1}) & \bfseries 0.46 & \bfseries 15\% & \bfseries 0.33 & \bfseries 57\% & 0.48 & \bfseries -4\% & 0.59 & 883\% & 184M \\
\bottomrule
\end{tabular}}
\caption{Evaluation of second-stage rerankers. Each metric score is given with its relative delta ($\Delta$) over the baseline. Inference time per query (sec/query) is measured on a GPU.}
\label{tab:reranker-evaluation}
\end{table*}

\section{LLM Prompts}
\label{sec:appendix-prompts}

\codeboxinput[label=lst:generator-prompt]{Prompt for question answering based on retrieved documents.}{figures/prompt-generator.txt}
\codeboxinput[label=lst:prompt-direct-answering]{Prompt for question answering without retrieval.}{figures/prompt-direct-answering.txt}
\codeboxinput[label=lst:prompt-classifier]{Prompt to classify a query into retrieval or no-retrieval.}{figures/prompt-classifier.txt}
\codeboxinput[label=lst:selectivity-prompt]{Prompt to detect non-answers for evaluation of selectivity.}{figures/prompt-selectivity.txt}

\section{Example FAQ Data}
\begin{table}[h!]
\small
\centering
\renewcommand{\arraystretch}{1.2}
\begin{tabular}{lp{.35\textwidth}p{.55\textwidth}}
\toprule
\# & \textbf{FAQ} & \textbf{Source}\\
\midrule
1 & Which Bachelor degree is needed to be admitted to the Masters of Data Science program? & ($D_1$) Master Data Science Admission Requirements;\newline($D_4$) Master Data Science Eligibility Assessment Process \\
2 & Which language certificates are accepted? & ($D_1$) Master Data Science Admission Requirements;\newline ($D_5$) Demonstrating English proficiency (C1);\newline ($D_6$) Demonstrating English proficiency (B2);\newline ($D_7$) Demonstrating German proficiency \\
3 & What are the application deadlines? & ($D_8$) Application Deadlines for Master Programs \\
\bottomrule
\end{tabular}
\caption{Example FAQ data. Each question is linked to answer-relevant documents in the knowledge base.}
\label{tab:faq-example}
\end{table}

\clearpage

\section{Example Evaluation Data}
\begin{table}[h!]
\small
\centering
\renewcommand{\arraystretch}{1.2}
\begin{tabular}{lp{.35\textwidth}p{.35\textwidth}p{.18\textwidth}}
\toprule
\# & \textbf{Question} & \textbf{Reference Answer} & \textbf{Source}\\
\midrule
1 & I am writing to ask whether me a Aerospace Engineering graduate can apply for the Master in Data Science? If yes will there be additional requirements? & Your bachelor degree needs to be in Data Science, Mathematics or Computer Science or comparable. Overall, at least 90 credit points (CP) must have been completed in in academic courses on Mathematics and Computer Science. Note that there is a list of specific courses that are required, and you need to cross-check whether you fulfil the requirements on those, too. Industry training or professional experience does not count as academic education. & ($D_1$) Master Data Science Admission Requirements \\\hline
2 & I have 2 questions: Is German language A2 required during online application? If I don't have proof of German language A2 level, Is it possible to be accepted? & You need to prove either English level C1, or German DSH-2 and English B1. Proof of language proficiency needs to be submitted with the application. & ($D_1$) Master Data Science Admission Requirements; ($D_2$) Application with a degree from abroad (first semester) \\\hline
3 & I aspire to take admission at your esteemed university for masters in data science for second semester in summer as I am pursuing masters in data science for first semester at a university in Germany currently for the winter semester. I wanted to know how to apply for second semester and the deadline for it. & In short, you need to obtain an advanced placement certificate from the departmental advisor of the Masters Data Science programme. You then and apply via uni-assist, indicating an application for the second semester. The process is explained in detail on the university website. & ($D_3$) Application with a degree from abroad (higher semester) \\\hline
4 & Can my application be checked before upload to see whether I am eligible. I do not want to pay the application fee if I would not be admitted to the program. & The admission committee cannot do any pre-application checks. You can check your transcript of records yourself and see whether it matches the admission criteria. & n/a \\\hline
5 & I am now in the process of preparing my papers for submission, but due to some circumstances, it may be a little delayed. In the end, everything will be ready before Deadline on February 23. My question is: Does delaying the application affect the chance of acceptance? Does the acceptance stage take place during the application period or does it not take place until after the deadline has passed for all applicants? & Applications are checked independent of each other, so your chance of admission does not depend on the time you submitted the documents. It is only checked whether the admission requirements are fulfilled. Offer letters will be sent as soon as the application is checked. & n/a \\\hline
6 & If I am unable to obtain my visa before the  semester begins, would it be possible to defer my admission to the next semester master data science? & You can transfer your admission to the next semester one time if the following conditions are met: (1.) You are not yet enrolled. (2) The study program is offered in the following semester (which will be for Master Data Science). (3) Admission Conditions in the Study Regulations have not changed (not planned for Master Data Science). You should then write an email to masters@uni-marburg.de in December / July and request the transfer. Then you will get a new letter of admission (after the check whether the above conditions are met). & n/a \\
\bottomrule
\end{tabular}
\caption{Sample questions from evaluation data. Each question has a manually written reference answer. Questions that are answerable with the knowledge base (\#1--3) also have ground-truth sources to evaluate retrieval. Unanswerable questions (\#4--6) are used to evaluate generator's ability to abstain.}
\label{tab:query-examples}
\end{table}

\clearpage

\section{Example Generations Reflecting Current Limitations}
\begin{figure}[h!]
\small\sffamily
\setlength{\currentparskip}{\parskip}
\begin{mdframed}
\setlength{\parskip}{0.3em}

\textbf{\color{NavyBlue} User:} I have an industrial engineering bachelor degree from iran butmy requirement courses are 50\% rekated to requirement of data science program can I be elighable?

\textbf{\color{orange} Marcel:} According to the Data Science (Master of Science) document, it is stated that "We require proof that you have completed a relevant Bachelor's degree in the fields of Data Science, Mathematics or Informatics or proof of a comparable domestic or foreign university degree qualifying you for a profession."

Additionally, it is required that "at least 72 credit points (CP) must have been earned in modules from the scientific discipline of Computer Science and at least 18 CP in modules from the scientific discipline of Mathematics."

However, it is not explicitly stated that the entire undergraduate degree program must be directly related to Data Science. Nevertheless, it is stated that "It is strongly recommended that students take responsibility for acquiring the skills from the above modules that they have not already acquired before commencing their studies."

Since you have mentioned that only 50\% of your requirement courses are related to the requirements of the Data Science program, it might be possible that you are eligible, but you would likely need to demonstrate that you have acquired the necessary skills in the areas where your undergraduate degree program was lacking.

You can contact Prof. Dr. Christin Seifert for content-related questions about the program, and Master's Coordination for formal questions about your application. Email: christin.seifert@uni-marburg.de Email: master@uni-marburg.de

\vspace{1em}

\textbf{\color{NavyBlue} User:} about elighbility assesment dose operations research is related to `Basic Real Analysis' or `Analysis I'

\textbf{\color{orange} Marcel:} Unfortunately, I do not have any information about Operations Research being directly related to 'Basic Real Analysis' or 'Analysis I' in the context of the Data Science program at Marburg University in the provided documents.

\vspace{1em}

\textbf{\color{NavyBlue} User:} so what is fundumentals of basic real analysis

\textbf{\color{orange} Marcel:} Based on the provided documents, the content of "Basic Real Analysis" is not explicitly stated. However, "Analysis I" and "Analysis II" are mentioned as alternatives to "Basic Real Analysis" in the prerequisites for some modules.

In the module "Stochastical Analysis", it is mentioned that either "Analysis I and Analysis II" or "Basic Real Analysis" is recommended. This suggests that "Basic Real Analysis" may cover similar topics as "Analysis I and Analysis II", but with a focus on real analysis.

In general, a course on real analysis typically covers topics such as:
\begin{itemize}[noitemsep,topsep=0pt]
    \item Real numbers and their properties
    \item Sequences and series of real numbers
    \item Continuity and differentiability of real-valued functions
    \item Integration of real-valued functions
    \item Metric spaces and topology
\end{itemize}
However, without more specific information, it is difficult to say exactly what topics are covered in the module "Basic Real Analysis" at Marburg University.

\vspace{1em}

\textbf{\color{NavyBlue} User:} what about Efficient Algorithms

\textbf{\color{orange} Marcel:} Efficient Algorithms is a module that is part of the Data Science program at Marburg University. It is mentioned as one of the recommended prerequisites for the Parameterized Algorithms module, and it is also mentioned as an example of a module that students may take as part of their studies in the Perspectives document.

According to the Online-Modulhandbuch, the "Efficient Algorithms" module (CS 566) is an advanced module that covers topics such as algorithmic techniques, greedy algorithms, dynamic programming, and space and time complexity. The module has a workload of 270 hours and is worth 9 credit points.

The qualification goals of the "Efficient Algorithms" module include:
\begin{itemize}[noitemsep,topsep=0pt]
    \item Applying skills in designing algorithms and knowledge of key design and analysis paradigms
    \item Using efficient data structures in algorithm design
    \item Analyzing algorithms with respect to correctness and effort
    \item Practicing scientific working methods (recognizing, formulating, solving problems, abstraction)
    \item Training to speak freely about scientific content
\end{itemize}
There are no explicit prerequisites for the "Efficient Algorithms" module, but the following modules are recommended:
\begin{itemize}[noitemsep,topsep=0pt]
    \item Algorithms and Data Structures
    \item Practical Informatics II: Data Structures and Algorithms for Pre-Service-Teachers
\end{itemize}
The module is assigned to the Computer Science department and can be attended by students in various study programs, including B.Sc. Data Science, B.Sc. Computer Science, M.Sc. Data Science, M.Sc. Computer Science, and others.
\end{mdframed}
\caption{Example where a user wants to determine if their academic background meets admission requirements.}
\label{fig:example-proactivity}
\end{figure}

\begin{figure}[ht!]
\small\sffamily
\setlength{\currentparskip}{\parskip}
\begin{mdframed}
\setlength{\parskip}{0.3em}
\textbf{\color{NavyBlue} User:} For Data Science can I submit my IELTS after deadline?

\textbf{\color{orange} Marcel:} Unfortunately, I do not have any information about the possibility to submit IELTS scores after the deadline for the Data Science Master's program.

\vspace{1em}

\textbf{\color{NavyBlue} User:} Can I submit my APS after deadline?

\textbf{\color{orange} Marcel:} Unfortunately, I do not have any information about the possibility to submit APS scores after the deadline for the Data Science Master's program
\end{mdframed}
\caption{Examples where the chatbot abstains from answering due to knowledge gaps.}
\label{fig:example-knowledge-gap}
\end{figure}

\begin{figure}[ht!]
\small\sffamily
\setlength{\currentparskip}{\parskip}
\begin{mdframed}
\setlength{\parskip}{0.3em}
\textbf{\color{NavyBlue} User:} Is a Thesis required for admission to msc data science?

\textbf{\color{orange} Marcel:} No, a thesis is not a requirement for admission to the Data Science Master's program. However, if a final thesis worth at least 10 credit points has already been completed in the fundamental university degree program, proof of this final thesis and its scope must be submitted as part of the eligibility assessment process.

\vspace{1em}

\textbf{\color{NavyBlue} User:} Is english required

\textbf{\color{orange} Marcel:} Yes, English proficiency of at least level B1 of the "Common European Framework of Reference for Languages" or English proficiency of at least level C1 is required, depending on the specific circumstances.

\vspace{1em}

\textbf{\color{NavyBlue} User:} What are the circumstances

\textbf{\color{orange} Marcel:} It appears that the circumstances refer to the situation in which initial applicants can move into an apartment straight away. According to the text, only under individual circumstances can initial applicants move into an apartment straight away, and in such cases, an early application is strongly advised.
\end{mdframed}
\caption{Example of a contextualization issue. The third query is a follow-up on the second query. Omission of the subject (``circumstances when B1 or C1 is required'') lead to retrieval/generation of unrelated information.}
\label{fig:example-contextualization}
\end{figure}

\begin{figure}[ht!]
\small\sffamily
\setlength{\currentparskip}{\parskip}
\begin{mdframed}
\setlength{\parskip}{0.3em}
\textbf{\color{NavyBlue} User:} Hi Marcel, what to study at uni marburg?

\textbf{\color{orange} Marcel:} You can find explanations of our degrees in the "Studying in Marburg" section, which may help you get a first impression of life in Marburg and the variety of subjects we offer.

\vspace{1em}

\textbf{\color{NavyBlue} User:} Can you show me a link to that?

\textbf{\color{orange} Marcel:} As a chatbot, I don't have the capability to share links. However, you can visit the University of Marburg's official website, and from there, you can navigate to the "Studying in Marburg" or the "Degree programmes" section. This will give you more information about the various courses they offer, including the MSc Data Science program that I can help you with.

Would you like to know more about the MSc Data Science program specifically?

\vspace{1em}

\textbf{\color{NavyBlue} User:} Why should I study Msc. DataScience?

\textbf{\color{orange} Marcel:} Unfortunately, I do not have any information about why you should study M.Sc. Data Science specifically, but you can find an overview of the program on the \href{https://www.uni-marburg.de/en/fb12/studying/degree-programs/m-sc-data-science/therefore-data-science}{M.Sc. Data Science web page}, which includes the required study skills and interests, as well as the content and structure of the program.

\end{mdframed}
\caption{Example of a link-seeking question. The generator cannot embed links to sources in the main text (second answer), unless some retrieved context happens to mention the link (third answer).}
\label{fig:example-link-seeking}
\end{figure}

\begin{figure}[t]
    \centering
    \includegraphics[width=.87\textwidth]{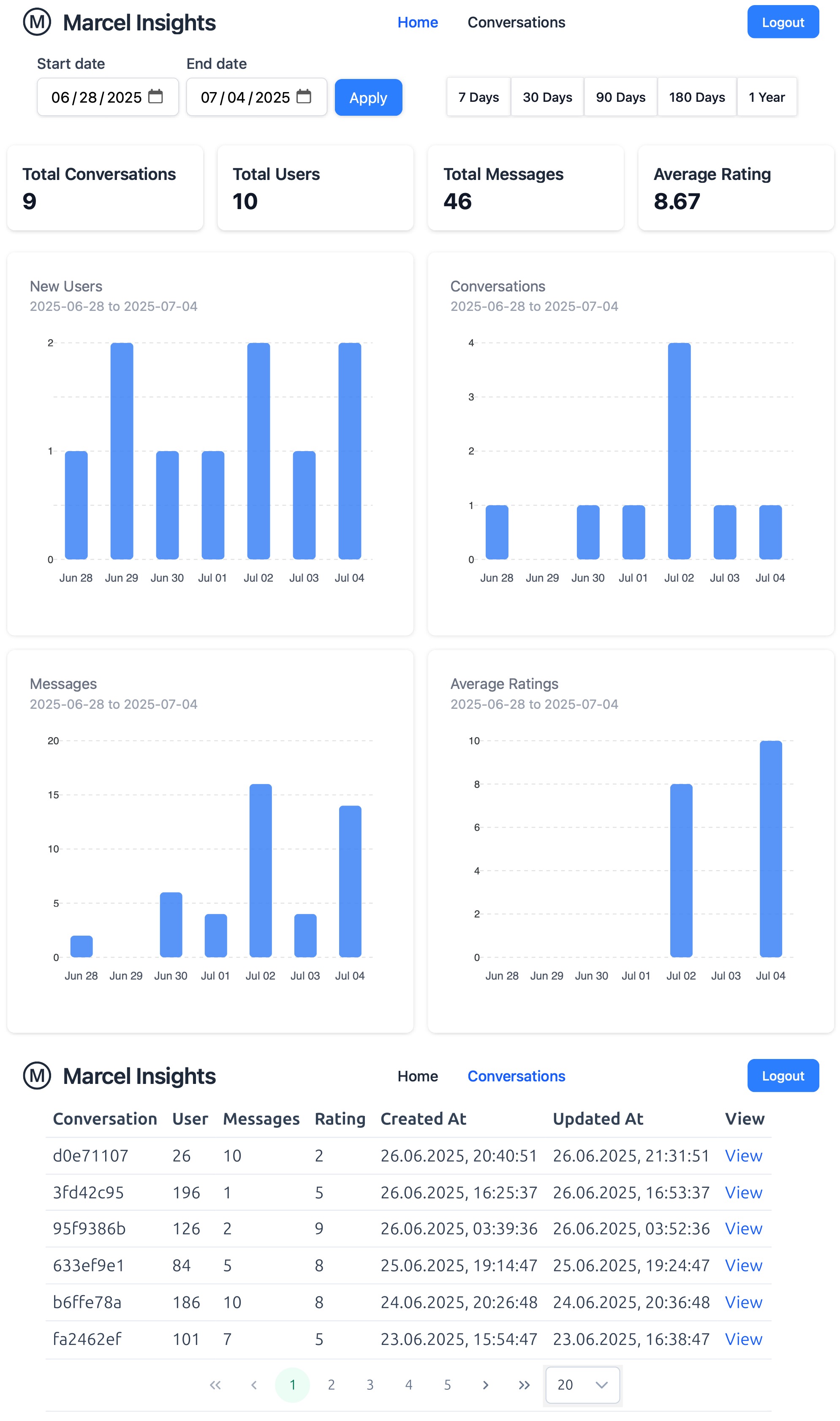}
    \caption{Overview of admin-facing UI which is only accessible to authorized users. \ul{Top:} A dashboard shows key usage statistics over a customizable time period. \ul{Bottom:} List of conversations with links for manual review.}
    \label{fig:admin-interface}
\end{figure}

\end{document}